\documentclass{article}

\usepackage[final]{neurips_data_2024}

\usepackage[utf8]{inputenc} 
\usepackage[T1]{fontenc}    
\usepackage{hyperref}       
\usepackage{url}            
\usepackage{booktabs}       
\usepackage{amsfonts}       
\usepackage{nicefrac}       
\usepackage{microtype}      
\usepackage{xcolor}         

\usepackage{enumitem}
\usepackage{multicol}
\usepackage{hyperref}
\usepackage{url}
\usepackage{wrapfig}
\usepackage{threeparttable}
\usepackage{bmpsize}
\usepackage{placeins} 

\usepackage{tikz}
\usepackage{tikz-cd}
\usepackage{quiver}

\usepackage{xspace}
\usepackage{enumitem}
\usepackage[framemethod=TikZ]{mdframed}
\usepackage{xspace}
\usepackage{amsmath,amssymb,amsfonts, amsthm}
\usepackage{cleveref}
\usepackage{subcaption}
\usepackage{inconsolata}
\usepackage{makecell, booktabs}
\PassOptionsToPackage{frozencache,cachedir=.}{minted}
\usepackage{minted}
\usepackage[normalem]{ulem}
\usepackage{appendix}
\usepackage{etoc}
\usepackage{comment}
\usepackage[symbol]{footmisc}
\usepackage{pgfplots}
\pgfplotsset{width=10cm,compat=1.15}

\usepackage[textsize=scriptsize]{todonotes}
\usepackage{xspace}
\usepackage{listings}

\usepackage{color}
\definecolor{keywordcolor}{rgb}{0.7, 0.1, 0.1}   
\definecolor{commentcolor}{rgb}{0.4, 0.4, 0.4}   
\definecolor{symbolcolor}{rgb}{0.0, 0.1, 0.6}    
\definecolor{sortcolor}{rgb}{0.1, 0.5, 0.1}      
\definecolor{errorcolor}{rgb}{1, 0, 0}           
\definecolor{stringcolor}{rgb}{0.5, 0.3, 0.2}    
\usepackage[textsize=scriptsize]{todonotes}

\DeclareUnicodeCharacter{211D}{$\mathbb{R}$}
\DeclareUnicodeCharacter{2260}{$\neq$}
\newcommand{\name}{\textsc{PutnamBench}\xspace}
\newcommand{\minif}{\textsc{MiniF2F}\xspace}
\newcommand{\fimo}{\textsc{Fimo}\xspace}

\title{\name: Evaluating Neural Theorem-Provers on the Putnam Mathematical Competition}

\author{%
  George Tsoukalas \\
  UT Austin\\
  \And
  Jasper Lee \\
  UT Austin \\
  \And
  John Jennings \\
  UT Austin \\
 \And
  Jimmy Xin \\
 UT Austin \\
 \AND
   Michelle Ding\\
   UT Austin \\
  \And 
  Michael Jennings\\
   UT Austin \\
   \And
   Amitayush Thakur\\
   UT Austin \\
   \And
   Swarat Chaudhuri \\
   UT Austin
}

\begin{document}

\maketitle

\begin{abstract}
We present \name, a new multi-language benchmark for evaluating the ability of neural theorem-provers to solve competition mathematics problems. \name consists of 1692 hand-constructed formalizations of 640 theorems sourced from the William Lowell Putnam Mathematical Competition, the premier undergraduate-level mathematics competition in North America. 
All the problems have formalizations in Lean 4 and Isabelle; a substantial subset also has Coq formalizations. \name requires significant problem-solving ability and proficiency in a broad range of topics taught in undergraduate mathematics courses. We use \name to evaluate several established neural and symbolic theorem-provers. 
These approaches can only solve a handful of the \name problems, establishing the benchmark as a difficult open challenge for research on neural theorem-proving. \name is available at \href{https://github.com/TrishulLab/PutnamBench}{https://github.com/trishullab/PutnamBench}.

\end{abstract}

\section{Introduction}
Automating mathematical reasoning is a longstanding goal in artificial intelligence \citep{newell1957empirical}.
A prominent line of work on the problem \citep{li2024survey} uses neural models to direct theorem-proving in formal frameworks like Lean 4 \citep{moura2021lean}, Isabelle \citep{wenzel2008isabelle}, and Coq \citep{Coq}. These frameworks can ``execute'' proofs like code and offer execution feedback, which simplifies the search for correct proofs. 

The design of quality benchmarks is a key challenge in this research area. The two most prominent competition-based benchmarks for neural theorem-proving are 
\minif \citep{zheng2021minif2f} and \fimo \citep{liu2023fimo}. The former formalizes a mix of problems from high-school level courses and mathematics competitions such as AIME, AMC, and IMO; the latter consists of a collection of IMO problems. Both benchmarks have limitations. For example, \minif contains many problems that can be immediately solved using an SMT solver, and \fimo  
only targets the Lean 3 framework, which is no longer actively maintained.

More generally, as large language models (LLMs) grow in importance as a tool for neural theorem-proving \citep{li2024survey}, preventing leakage between pretraining sets and evaluation sets is more important than ever. This makes the continued supply of new benchmarks an important goal. 

In this paper, we respond to this challenge with \name, a new hand-curated, multi-langauge benchmark for neural theorem-provers. \name includes 1692 formalizations of 640 problems from the William Lowell Putnam Mathematical Competition, the premier college-level mathematics competition in North America.\footnote{
\name is available at \href{https://github.com/TrishulLab/PutnamBench}{https://github.com/trishullab/PutnamBench}.} 
All our problems have Lean 4 \citep{moura2021lean} and Isabelle \citep{wenzel2008isabelle} formalizations; a substantial fraction have formalizations in Coq \citep{Coq} as well. The formalizations are all manually constructed and have been carefully debugged. The benchmark also includes the original English-language problem statements with permission from the Mathematical Association of America, which organizes the Putnam competition.

One key benefit of \name is that Putnam competition problems require a broad base of mathematical knowledge and skills. Because they target undergraduate students, they cover topics such as analysis and abstract algebra that do not appear in the International Mathematical Olympiad (IMO). At the same time, success in the two competitions is correlated --- top performers on the Putnam competition are often former IMO medalists as well. Hence, \name is well-aligned with the IMO Grand Challenge \citep{imograndchallengeGrandChallenge} and the AI Mathematical Olympiad \citep{aimoprizeAIMOPrize}, the latter of which offers a \$10M prize fund for developing a system that can win a gold medal at the IMO. 

Another advantage is that \name supports multiple proof assistants. Lean 4, Coq, and Isabelle are currently the three most popular formal proof languages. However, theorem-proving benchmarks typically only contain problems in a strict subset of these languages --- for example, \minif \citep{zheng2021minif2f} does not include Coq problems, and \fimo \citep{liu2023fimo} only targets Lean. \name is the first mathematics-competition benchmark to include problems in all three languages.

We use \name to evaluate several neural and symbolic approaches: Draft-Sketch-Prove \citep{jiang2022draft}, {\sc Copra} \citep{thakur2024incontext}, GPT-4, Sledgehammer \citep{paulsonsledgehammer}, and Coqhammer \citep{czajka2018hammer}. Collectively, these methods can only solve a handful of the \name problems, establishing \name as a hard open challenge for the neural theorem-proving community. 

\section{Background}
\newcommand{\state}{\mathcal{O}\xspace}

\paragraph{Formal Theorem-Proving.}

\begin{wrapfigure}{r}{0.5\textwidth}
\vspace{-0.8in}
\begin{mdframed}[roundcorner=10pt]
\begin{lstlisting}
theorem putnam_1988_b1 :
∀ a ≥ 2, ∀ b ≥ 2, ∃ x y z : ℤ,
x > 0 ∧ y > 0 ∧ z > 0 ∧ 
a * b = x * y + x * z + y * z + 1 := by
    intro a ha b hb
    use a - 1, b - 1, 1
    constructor
    linarith
    constructor
    linarith
    constructor
    linarith
    ring
\end{lstlisting}
\end{mdframed}
\caption{A formalization of Putnam 1988 B1 in Lean 4, which asserts that for all integers $a,b \geq 2$, there are positive integers $x,y,z$ such that $ab = xy + xz + yz + 1$. The formal proof begins by introducing all relevant variables and hypotheses with \texttt{intro}, then indicating the choice of $x,y,z$ with \texttt{use}, and afterwards proving all goals using the automated tactics \texttt{linarith} and \texttt{ring}. This proof was discovered through a few-shot invocation of GPT-4.} \label{fig:lean-example} 
\vspace{-0.3in}
\end{wrapfigure}

Formal proof frameworks like Lean 4 \citep{moura2021lean}, Coq \citep{Coq}, and Isabelle \citep{wenzel2008isabelle} allow users to write machine-verifiable proofs of mathematical theorems.
To create such a proof, one first uses a framework-specific language to formally state the target theorem. The mathematical objects referenced in the theorem can be imported from an existing repository or defined by the user. 
During the proof process, the proof framework maintains a \emph{state} that includes information about the parts of the proof that remain to be completed. One can change this state by executing a 
\emph{proof step}. The user's goal is to write a sequence of proof steps (in the framework's language) that changes the proof state to a special state ``QED'' in which there are no unmet proof obligations.

\Cref{fig:lean-example} illustrates a theorem and proof in the Lean 4 framework. 

\paragraph{The Putnam Competition.}

The William Lowell Putnam Mathematical \citep{maaWilliamLowell}, organized by the Mathematical Association of America (MAA), is the premier collegiate mathematics competition in North America. Thousands of undergraduate students from universities across the United States and Canada take the exam each year. The competition comprises two 3-hour-long sessions of six problems each, presented in approximately ascending order of difficulty within each session. While some problems require competitors to furnish a concrete solution (such as a number, a set, or the truth value of a given statement), all problems require a natural-language proof of correctness. The contest draws from a wide variety of topics in the undergraduate curriculum, often using instances of ideas from research-level mathematics.

\section{\textsc{PutnamBench}}
\label{sec:benchmark}
\begin{table*}[t]
  \centering

  \label{tab:benchmark-comparison}
  \begin{tabular}{l*{6}{c}}
    \toprule
    \textbf{Benchmark} & $\#$ & \textbf{Natural Language} & \textbf{Lean} & \textbf{Isabelle} & \textbf{Coq} & \textbf{Factored Solution} \\
    \midrule
    \textsc{miniF2F} & 488 & $\checkmark$ & $\checkmark$\footref{note:fimo-lean3} & $\checkmark$ & & \\
    \textsc{ProofNet} & 371 & $\checkmark$ &$\checkmark$\footref{note:fimo-lean3} & & & N/A \\
    \textsc{Fimo} & 149 & $\checkmark$ & $\checkmark$\footref{note:fimo-lean3} & & &  \\
    \textsc{PutnamBench} & 640 & $\checkmark$ & $\checkmark$ & $\checkmark$ & $\checkmark$ & $\checkmark$ \\ 
    \bottomrule
  \end{tabular}
    \caption{Comparison of existing formal theorem proving evaluation benchmarks. \name exceeds prior benchmarks by providing support for all of Lean 4, Isabelle, and Coq, on a set of difficult competition problems using undergraduate-level mathematics. For problems requiring a numerical solution in addition to a proof, we factor the solution out of the theorem statement.}
\end{table*}

\begin{wraptable}{r}{2in}
\vspace{-0.2in}
\begin{tabular}{lc}
\textbf{Category}      & \textbf{Total Quantity} \\ \hline
Algebra                & 253                     \\ 
Analysis               & 226                     \\ 
Number Theory          & 107                      \\ 
Geometry               & 68                      \\ 
Linear Algebra         & 51                      \\ 
Abstract Algebra       & 28                      \\ 
Combinatorics          & 26                       \\ 
Probability             & 9                      \\ 
Set Theory            & 8                       \\  \hline
\end{tabular}
\label{tab:category-split}
\caption{Quantity by domain of \name problems. Our formalizations generally reflect the variety of Putnam problems, though we can only formalize few geometry and probability problems due to limited support for these topics in the respective mathematical libraries.}
\vspace{-0.2in}
\end{wraptable}

\name is a multi-language evaluation benchmark consisting of formalized problems from the Putnam competition. 
\name is a manually produced benchmark, including 640 formalizations in Lean 4 and Isabelle, and 412 formalizations in Coq. In aggregate, \name contains 1692 formalizations of Putnam competition problems. We also incorporate the informal statements and numerical solutions where applicable. 

Now we elaborate on the main features of \name. 

\smallskip 
\textbf{Diversity and Breadth.} 
Compared to \minif \citep{zheng2021minif2f} and \fimo \citep{liu2023fimo}, which generally rely on high-school mathematics, \textsc{PutnamBench} incorporates a wider variety of problems which require definitions of the standard undergraduate mathematics curriculum. The \textsc{ProofNet} benchmark \citep{azerbayev2023proofnet} also sources problems from the undergraduate curriculum, but these problems are generally from standard textbooks as opposed to mathematical competitions. Putnam problems often require definitions from multiple fields, which standard textbooks do not necessarily target. Formalizations in \textsc{PutnamBench} include concepts from a wide range of mathematical fields, including:
(i) \emph{\textbf{Analysis}}: Limits, integrals, derivatives, continuity; 
(ii) \emph{\textbf{Linear Algebra}}: Matrices, determinants, fields; 
(iii) \emph{\textbf{Abstract Algebra}}: Rings, groups, magmas, permutations; 
(iv) \emph{\textbf{Algebra}}: Polynomials, inequalities, algebraic expressions; 
(v) \emph{\textbf{Number Theory}}: Primes, irrationality, base representations, divisors, palindromes; 
(vi) \emph{\textbf{Geometry}}: Polygons, point sets, line intersections, Euclidean distance; 
(vii) \emph{\textbf{Set Theory \& Combinatorics}}: Countability, power sets, discrete structures, games.
\smallskip 

\textbf{Multiple Languages.} \name contains formalizations of Putnam problems in Lean 4, Isabelle, and Coq. The formalizations also include concepts defined in each proof assistant's mathematical repositories --- notably, Mathlib, the HOL standard library, and Coquelicot  (among various Coq repositories). To the best of our knowledge, \name is the first undergraduate-level competition benchmark for each of these languages. Furthermore, we are the first to produce a human mathematics competition-style evaluation benchmark for Coq. 

We hope that this contribution can enable Coq practitioners access to the rapidly-growing field of machine learning for mathematics.

Generally, the formalizations of the problems are aligned in their structure, including hypothesis naming and framing. Differences may arise according to the underlying foundations of each language. We also note that the pre-defined mathematical theory in each language differs, which can sometimes lead to difficulties formalizing certain problems. 

Compared to the prior benchmarks \minif, \fimo, and \textsc{ProofNet}, \name is the first to support Lean 4 on initial release \footnote{\label{note:fimo-lean3}\small{\minif, \fimo, and \textsc{ProofNet} were originally released using Lean 3, and \minif and \fimo have been updated to include Lean 4 formalizations following community efforts. \citep{azerbayev2023proofnet, githubGitHubRahul3613ProofNetlean4}. To the best of our knowledge, no open-sourced Lean 4 version of FIMO currently exists.}}.

\smallskip 
\textbf{Factored Solutions.} Roughly 60\% of Putnam problems, in their natural language form, require exhibiting a (closed-form) solution along with a proof of its correctness. Such problems do not assert propositions, and hence are not immediately formalizable as they are not directly the statement of a theorem. Prior benchmarks such as \minif \citep{zheng2021minif2f} sidestep this issue by rewording the problem statement to ask for a proof that the solution satisfies the constraints of the problem. However, this reduction diminishes the overall difficulty of the problem, as producing a solution can constitute the majority of the difficulty. To address this issue, we factor out solutions of such problems from the formalized theorem statement. We include an example in \Cref{fig:factored-solution-example}. In this way, we provide two tasks for neural theorem proving:

\begin{itemize}[leftmargin=0.2in]
\item \textbf{Task 1:} Given the theorem statement, first identify the (closed-form) solution, and then provide a proof of correctness by rewriting the solution into the theorem statement.

\item \textbf{Task 2:} Given the theorem statement and solution, produce a proof of its correctness. This task aligns with the current benchmarks.
\end{itemize}

We note that the process of producing the numerical solution may be highly correlated with the proof of its correctness. In this way, our formalizations can reflect the true difficulty of the informal problem statement.

\begin{wrapfigure}{r}{0.55\textwidth}
\vspace{-0.22in}
\begin{mdframed}[roundcorner=10pt]
\textbf{Putnam 2008 B5.} Find all continuously differentiable functions $f : \mathbb{R} \to \mathbb{R}$ such that for every rational number $q$, the number $f(q)$ is rational and has the same denominator as $q$.
\end{mdframed}
\begin{mdframed}[roundcorner=10pt]
\begin{lstlisting}
abbrev solution : Set (ℝ → ℝ) := 
    {fun (x : ℝ) => x + n | n : ℤ} ∪ 
    {fun (x : ℝ) => -x + n | n : ℤ}
theorem putnam_2008_b5 
(fqsat : (ℝ → ℝ) → ℚ → Prop)
(hfqsat : ∀ f q, fqsat f q ↔ 
    ContDiff ℝ 1 f ∧ 
    (∃ p : ℚ, p = f q ∧ p.den = q.den)) :
∀ f : (ℝ → ℝ), (∀ q : ℚ, fqsat f q) 
↔ f ∈ solution :=
\end{lstlisting}
\end{mdframed}
\caption{A formalization of Putnam 2008 B5 in Lean 4. As the problem requires exhibiting the set of functions $f$ satisfying the specified conditions, it is not directly the statement of a theorem. We formalize the problem by instantiating a variable ``solution'' outside of the theorem statement. In this way, a model can either provide its own candidate, or use the correct solution we provide and attempt to produce a proof of correctness. Benchmarks such as \minif and \fimo only include formalizations with the solution written into the theorem statement.}
\label{fig:factored-solution-example} 
\vspace{-0.2in}
\end{wrapfigure}
\smallskip 

\textbf{Formalization effort and challenges.} We hand-crafted our benchmark over the course of several months as a team of two doctoral and five undergraduate students with prior experience in university mathematics, computer science, and formal proof assistants. We found that the average time-to-formalize a single problem in one language was roughly 25 minutes. 
Each formalization was verified by a second person at least once, and we measured that the verification  of a single formalization took between 10 minutes, on average. We acknowledge that the time-to-formalize we report is higher than that of \minif; we believe this is largely due to the increased complexity of the Putnam problems, which oftentimes require definitions we must locate in each language's respective mathematical libraries. 

We first produced formalizations in Lean 4, and then proceeded with our formalization effort in Isabelle and then Coq. Due to differences in the underlying foundations of each language, we found that formalizations in one language sometimes do not directly transfer to another; for example, Isabelle does not have a subtyping mechanism, which we made extensive use of in Lean 4. Formalizations in Coq rely on a number of mathematics repositories. Predominantly, we rely on MathComp and MathComp-Analysis \citep{githubGitHubMathcompmathcomp, githubGitHubMathcompanalysis}, but also make us of Stdlib, Stdpp, Coquelicot, GeoCoq, and Coqtail \citep{githubGitHubTherycoquelicot, githubGitHubGeoCoqGeoCoq, githubGitHubWhonoreCoqtail}.

Some problems are not naturally amenable to formalization --- for example, we found that while formalizing problems involving probabilities is possible, such formalizations often require heavy probability theory.
Similarly, support for problems involving Euclidean geometry varies across languages; in particular, Lean 4 does not yet have a sufficiently extensive library to make most geometry problems formalizable. By contrast, Coq has an extensive geometry repository called GeoCoq, which we utilize for our Coq formalizations.

\begin{wrapfigure}{r}{0.6\textwidth}
\vspace{-0.05in}
\begin{mdframed}[roundcorner=10pt]
\begin{lstlisting}
(a) theorem putnam_2006_b2
(n : ℕ)
(npos : n > 0)
(X : Finset ℝ)
(hXcard : X.card = n)
: (∃ S ⊆ X, S ≠ ∅ ∧ ∃ m : ℤ, 
    |m + ∑ s in S, s| ≤ 1 / (n + 1))
\end{lstlisting}
\end{mdframed}
\begin{mdframed}[roundcorner=10pt]
\begin{lstlisting}
(b) theorem putnam_2006_b2:
fixes n :: nat
and X :: "real set"
assumes npos: "n > 0"
and hXcard: "finite X ∧ card X = n"
shows "∃ S ⊆ X. (S ≠ {}) ∧ (∃ m :: int. 
    ¦m + (∑ s ∈ S. s)¦ ≤ 1 / (n + 1))"
\end{lstlisting}
\end{mdframed}
\begin{mdframed}[roundcorner=10pt]
\begin{lstlisting}
(c) Theorem putnam_2006_b2
(n : nat)
(hn : gt n 0)
(X : seq R)
(hX : uniq X /\ size X = n)
: exists S : seq R, 
    subseq S X /\
    size S <> 0%nat /\
    exists m : int, 
    `|m%:~R + \sum_(s <- S) s| <= 1 / (n%:R + 1).
\end{lstlisting}
\end{mdframed}
\caption{Formalizations of Putnam 2006 B2 in (a) Lean 4, (b) Isabelle, (c) Coq. Putnam 2006 B2 asserts that given a finite subset $X \subseteq \mathbb{R}$ with $|X| = n > 0$, there is a nonempty subset $S \subseteq X$ and an $m \in \mathbb{Z}$ such that $|m + \sum_{s \in S} s| \leq \frac{1}{n+1}$.} \label{fig:multilingual-example}
\vspace{-0.3in}
\end{wrapfigure}

\smallskip 
\textbf{Dataset Contamination.} Our benchmark is unique compared to informal benchmarks such as MATH \citep{hendrycks2021measuring} and GSM8K \citep{cobbe2021training} in the sense that the target output \emph{has never been produced}, hence avoiding direct contamination. To the best of our knowledge, we are the first to provide formalizations of a large collection of Putnam problems in any of Lean, Isabelle, and Coq. Since writing a formal proof requires the formal theorem statement, it is highly unlikely any possible formal proof has been written for any of our problems. We performed a thorough investigation of formal mathematics repositories for each language for confirmation, finding no aligned theorems and proofs from the Putnam Competition. 
We do not include any of the formal proofs in our benchmark.

Furthermore, any proofs found by automated methods in our evaluations are not included and are only mentioned in this article. Indirect contamination can occur through transfer from training on the informal proofs, though producing proofs in formal proof environments still presents a major difficulty for all current neural methods, as we find in \Cref{sec:eval}.

\smallskip 
\textbf{Licensing and Rules of Engagement.} \textsc{PutnamBench} is available under an Apache 2.0 license for Lean 4 and Isabelle, and under an MIT license for Coq. We align the licenses with those of the repositories we use for each language. With permission from the MAA, we include the informal statements as sourced from the competition \citep{alexanderson1985william, kedlaya2002william, kedlaya2020william}. We host a public leaderboard at \href{https://trishullab.github.io/PutnamBench/}{https://trishullab.github.io/PutnamBench/} and will readily accept evaluation results from future works.

\section{Experimental Evaluation}\label{sec:eval}
To understand the challenges that \name poses for state-of-the-art theorem-proving approaches, we attempt to solve its problems using a suite of such approaches. 
Given the relative lack of tailored systems for multi-language theorem-proving, we run evaluations for each language separately. Any method that is evaluated on multiple languages is based on off-the-shelf foundation models. 

\smallskip 
\textbf{Metrics.} Our evaluation is based on the $pass@n$ \citep{lample2022hypertree} metric. This metric measures a prover's ability to produce a successful proof, as determined by the formal proof environment, given a budget of $n$ \emph{proof attempts}. In search-based methods  \citep{thakur2024incontext}, each proof attempt involves a distinct search that can query a neural model multiple times. 

\smallskip 
\textbf{Models.}
For each of the languages, we perform evaluations using GPT-4 \citep{openai2023gpt4} \footnote{We use GPT-4o for all our evaluations}, a highly capable foundation model. We run evaluations using in-context learning, appending several examples of successful proofs of simple theorems in each language. For evaluations with Lean 4 approaches, we note that many approaches have targeted Lean 3, which is not backward-compatible and no longer actively maintained. 
We evaluate {\sc Copra} \citep{thakur2024incontext} on \name, modifying the prompt examples of {\sc Copra} to enable search in Lean 4. Furthermore, we run evaluations  LeanDojo's retrieval-augmented prover {\sc ReProver}, a finetuned model designed to utilize incorporate retrieved lemmas as part of the proof search. We also include evaluations with the retrieval component held out.

For our Isabelle experiments, we run evaluations of Draft, Sketch, and Prove (DSP) \citep{jiang2022draft} using GPT-4 as the underlying foundation model, noting that many further works for theorem-proving in Isabelle have extended on the DSP pipeline as we mention in \Cref{sec:related-works}. We also run evaluations using stand-alone invocations to Sledgehammer, a powerful symbolic automation tool in Isabelle that relies on calls to external SMT solvers. 

As for our Coq experiments, prior neural approaches for Coq have mostly targeted software verification tasks, as opposed to competition mathematics. 
As a result, our Coq experiments use {\sc Copra}, which also supports theorem-proving in Coq. We evaluate using the Tactician \citep{Blaauwbroek_2020} platform with the locality sensitive hashing model configuration. We also run evaluations using CoqHammer \citep{czajka2018hammer}, a tool similar to Isabelle's Sledgehammer, which makes calls to external constraint solvers.

\subsection{Results}

\begin{table}[t]
\vspace{-0.05in}
    \begin{minipage}{.3\linewidth}
      \centering
      \caption*{\name : Lean}
        \begin{tabular}{@{}lc@{}}
            \toprule
            Method & Success Rate \\ \midrule
            GPT-4 & 1/640 \\
            COPRA & 1/640 \\
            ReProver ($+$r) & 0/640 \\
            ReProver ($-$r) & 0/640 \\
            \bottomrule
        \end{tabular}
    \end{minipage}%
    \hspace{1em} 
    \begin{minipage}{.3\linewidth}
      \centering
      \caption*{\name : Isabelle}
        \begin{tabular}{@{}lc@{}}
            \toprule
            Method & Success Rate \\ \midrule
            GPT-4 & 1/640 \\
            DSP & 4/640 \\
            Sledgehammer & 3/640 \\
            \bottomrule
        \end{tabular}
    \end{minipage} 
    \hspace{1em} 
    \begin{minipage}{.3\linewidth}
      \centering
      \caption*{\name : Coq}
        \begin{tabular}{@{}lc@{}}
            \toprule
            Method & Success Rate \\ \midrule
            GPT-4 & 1/412 \\
            COPRA & 1/412 \\
            Tactician & 0/412 \\
            CoqHammer & 0/412 \\
            \bottomrule
        \end{tabular}
    \end{minipage}
    \vspace{0.1in}
    \caption{Results of evaluations on \name in each language. We find that all tested methodologies perform poorly, solving at most a handful of problems. Notably, the only problem solved in both Lean and Coq is Putnam 1988 B1, which is not solved by any method in Isabelle. ReProver, our finetuned baseline for Lean, is unable to solve any problems with or without retrieval. Symbolic automation proves to be powerful in Isabelle, with Sledgehammer solving the most problems than GPT4 alone. DSP generates four successful proofs, two of which cannot be generated by Sledgehammer alone.}
\vspace{-0.25in}
\end{table}
\paragraph{Lean 4.} We prompt GPT-4 in a $pass@10$, setting temperature $T = 0.7$ and using several examples of simple theorems and proofs, to generate a proof for each problem. The result of this experiment yields a single successful proof across all 640 Lean formalizations. The problem (Putnam 1988 B1) and the generated proof are given in \Cref{fig:lean-example}. In particular, Putnam 1988 B1 is solved on the first of 10 attempts. An example of a failure mode of GPT-4 is given in \Cref{fig:appendix-copra-premise-failure}.

We also run evaluations with COPRA, using their default hyperparameters for search, performing a $pass@1$, and allowing 60 queries to GPT-4. However, since COPRA was originally designed for interaction with Lean 3, we make a small modification to its system prompt to enable search in Lean 4. The result of the step-wise proof search over all Lean 4 formalizations yields a correct proof to one problem (1988 B1). We find that backtracking in the search was not required for this proof, which was 10 lines long and was found at the 10th query. It is possible that affording COPRA further queries to GPT-4 can yield more successful proofs, though it is not yet feasible to perform such an experiment due to the cost of queries to GPT-4.

We found that, by default, GPT-4 produces proofs using Lean 3 syntax, which is not compatible with Lean 4. Even when directed to produce outputs in Lean 4, GPT-4 typically continues to produce outputs in Lean 3. Our prompt, which we include in \Cref{fig:gpt-prompt}, elucidates some design differences in Lean 4 to better enforce compliance with the Lean 4 syntax. However, we noticed many examples where GPT-4 continues to output terms in Lean 3 syntax. One such example is given in \Cref{fig:appendix-copra-lean3-failure}.

We run {\sc ReProver} using the standard search parameters used in LeanDojo \citep{yang2023leandojo}. Our evaluation yields no successfully proven problems, with and without the 
inclusion of the retrieval module. We believe that Putnam 1988 B1, which the other methods solve, is not solved by {\sc ReProver} as it requires an understanding that the choice of $x,y,z=1,a-1,b-1$ will eventually satisfy the conditions of the goal after simplification. Smaller models, like the one driving {\sc ReProver}'s search, may not be as readily capable of such understanding.

\paragraph{Isabelle.}
We run GPT-4 using the same configuration, with modified prompts for Isabelle, on our Isabelle formalizations. We find that GPT-4 can produce a single successful proof to Putnam 1986 B1, a geometric problem stated algebraically. We include the statement and its proof as generated by GPT-4 in \Cref{fig:geometric-problem}.

\begin{wrapfigure}{r}{0.6\textwidth}
\vspace{-0.25in}
\begin{mdframed}[roundcorner=10pt]
\textbf{Putnam 2001 A1.} Consider a set $S$ and a binary operation $\star$, i.e., for each $a,b \in S$, $a \star b \in S$. Assume $(a \star b) \star a = b$ for all $a,b \in S$. Prove that $a \star (b \star a) = b$ for all $a,b \in S$.
\end{mdframed}
\begin{mdframed}[roundcorner=10pt]
\begin{lstlisting}
theorem putnam_2001_a1:
  fixes op :: "'a ⇒ 'a ⇒ 'a"
  assumes hop : "∀a b :: 'a. 
    op (op a b) a = b"
  shows "∀a b :: 'a. op a (op b a) = b"
proof -  
  {
    fix a b :: 'a     
    have "op (op a (op b a)) a = op b a" using hop by simp    
    then have "op a (op b a) = b" using hop by metis  
  }   
  then show ?thesis by simp 
qed
\end{lstlisting}
\end{mdframed}
\caption{A formalization of Putnam 2001 A1 in Isabelle and the corresponding proof discovered by our evaluation with DSP. Sledgehammer alone can also produce a successful proof to this theorem. } 
\label{fig:isabelle-example} 
\vspace{-0.2in}
\end{wrapfigure}

DSP represents a neurosymbolic methodology which has seen significant application for theorem-proving in \minif. We run DSP with $pass@10$, using temperature $T = 0.1$ and GPT-4 as the underlying language model. Our evaluation yields four successful proofs: of Putnam 2001 A1 and 1971 B1, two problems involving magmas (sets with a binary operation), one of Putnam 1995 A1, a problem involving a closed-under-multiplication subset of the reals, and Putnam 1986 B1. In particular, Putnam 1995 A1 and 1986 B1 cannot be solved by Sledgehammer alone. The generated proof of Putnam 1995 A1 is included in \Cref{fig:isabelle-example}.

We run a baseline using Sledgehammer, a powerful automation tool in Isabelle which makes calls to external SMT solvers to prove a given goal. With a set timeout of $t = 120$ seconds, we run Sledgehammer on each Isabelle formalization. The result of this evaluation is 3 successfully proven problems: Putnam 1971 B1, 2001 A1, and 2012 A2. Notably, all of these problems are statements about sets with binary operations. We include the statements of 1971 B1 and 2012 A2 in \Cref{fig:isabelle-sledgehammer-others}.

\paragraph{Coq.} We run GPT-4 with a Coq-based prompt on our Coq formalizations using the same configuration as in Lean and Isabelle. The result of the experiment is 1 solved problem, namely Putnam 1988 B1, which was also solved in Lean 4. The proof, which we include in \Cref{fig:coq-proof}, generally follows the same structure as the proof in Lean.

An evaluation with COPRA, in a $pass@1$-with-$60$-queries and $T = 0.0$ also yields a successful proof only for Putnam 1988 B1 which we include in \Cref{fig:coq-proof}. In this case, backtracking was crucial for proof search on this problem. The crucial step in 1988 B1 is the choice of $x,y,z$ once $a$ and $b$ have been introduced. Initially, COPRA predicts the erroneous choice $x, y, z = 1, 1, ab-1$ and eventually reverts this choice using backtracking. Afterwards, COPRA predicts a correct choice $x, y,z = 1, a-1, b-1$ and proceeds with the proof. 

We run Tactician using the locality sensitive hashing model with a timeout of $t = 600s$ per problem. Our evaluation yields no successfully proven problems. While showing favorable performance on theorems drawn from Coq's standard library \citep{zhang2021onlinemachinelearningtechniques}, such methodologies do not as of yet scale to challenging olympiad-style problems. 

We run CoqHammer with 8 parallel threads using an ATP timeout of 100 seconds, proof reconstruction timeout of 15 seconds, and sauto timeout of 5 seconds, for a total of 120 seconds allocated for each formalization. The evaluation yields no successful proofs --- indicating that symbolic tools in Coq are not yet capable of handling \name problems. It is not surprising that CoqHammer does not match the performance of Sledgehammer even though they rely on the same external solvers. The underlying logical system of Coq is more complex than that of Isabelle and is hence less amenable to automation.
\subsection{General Analysis}
Aggregating over all experiments performed in all languages, we find that a total of 6 problems in \name are successfully proven. A majority of these come from evaluations in Isabelle, particularly with strong contributions from Sledgehammer. Sledgehammer can solve all three problems involving magmas which appear in our benchmark but fails to produce successful proofs for any other formalization. DSP solves an additional two problems and relies heavily on Sledgehammer to fill in the proofs of intermediate steps. The single problem solved in Lean and Coq also makes use of automated tactics like \texttt{linarith} and \texttt{lia}, and requires only a single crucial step. 

Hence, we find that a few \name problems are not entirely intractable using current methods. However, anecdotally, these problems are among the easiest ever included in the Putnam competition. All admit a very short natural language proof and do not require reasoning about particularly complicated objects. We believe that significant advancements in automated mathematical reasoning are required to make progress on \name.

\section{Related Work}\label{sec:related-works}
\textbf{Formal Benchmarks.}
Several evaluation benchmarks for formal mathematics have been developed in recent years. \minif \citep{zheng2021minif2f} is a formal-to-formal benchmark of competition problems, sourced from high school competitions such as the AMC, AIME, and IMO. \minif is a multi-language benchmark, comprising of 488 problems each formalized in Lean 3, Metamath, Isabelle and HOL Light. We chose not to include formalizations in Metamath and HOL Light as they have not been the focus of attention for neural theorem-proving. A similar competition-style benchmark is FIMO \citep{liu2023fimo}, which contains 149 Lean 3 formalizations of IMO shortlist problems produced using a back-translation procedure with GPT-4. The automatically-generated formalizations are then manually verified. Both benchmarks are designed to measure \emph{certifying} the solution to the informal problem statement when one exists. \citet{compfiles} is a collection of 171 Lean 4 formalizations of competition problems, predominantly from the IMO and USAMO, often accompanied by a formal proof, which has not seen use in benchmarking automated theorem-provers. ProofNet \citep{azerbayev2023proofnet} introduced a benchmark of 371 exercises, formalized in Lean 3, from standard textbooks in the undergraduate mathematics curriculum. While largely not competition-based, problems in ProofNet draw from a broader library of concepts than miniF2F and FIMO, which rely only on high-school mathematics. LeanDojo \citep{yang2023leandojo} introduces a dataset of formal mathematics and proofs derived from Lean's mathlib library \citep{mathlib}, and trains a retrieval-augmented model towards generating proofs on their held-out test set. ProverBot9001 \citep{sanchez2020generating} introduced a dataset for theorems and proofs written in Coq derived from CompCert \citep{compcert}, a formally verified C compiler. PISA \citep{jiang2021lisa} is a dataset derived from Isabelle's Archive of Formal Proofs \citep{ArchiveFormalProofs}, which contains theorems and proofs from general mathematics as opposed to specifically competition problems.

\smallskip 
\textbf{Informal Benchmarks.}
There are also several popular benchmarks for informal (natural-language) mathematical reasoning. MATH \citep{hendrycks2021measuring} is a collection of 12,500 mathematics problems, in natural language only, sourced from various high school competitions additionally supplied with step-by-step informal proofs. GSM8K \citep{cobbe2021training} is a collection of 8,500 grade school mathematics problems, intended to benchmark natural language reasoning for mathematics-style problems. While benefiting from the abundance of natural language data, these benchmarks fall short, since in natural language, there is no automatic mechanism for certifiable verification of the reasoning path which yielded the numerical answer. For this reason, metrics for success on these benchmarks usually rely on exact-answer match, because verifying reasoning paths is imprecise and is best done by human experts. By contrast, theorem proving in formal proof assistants comes with a high-confidence signal for correctness of the reasoning path, or \emph{proof}, of a theorem. 

\smallskip 
\textbf{Methods for Formal Theorem-Proving.}
Significant effort has been spent on developing automatic theorem-provers for formal mathematics \citep{li2024survey}. Most recent efforts train a neural module to perform proof-step prediction, which is then wrapped in a search mechanism to locate a valid proof. GPT-$f$ \citep{polu2020generative} trains a transformer-based architecture on data derived from the Metamath library \citep{megill2019metamath} for proof synthesis. PACT expands on GPT-$f$ by incorporating auxiliary training tasks for the neural module towards theorem-proving in Lean 3. FMSCL \citep{polu2022formal} alternates proof-search and training to finetune their neural model based on proofs found during search. HTPS \citep{lample2022hypertree} uses a transformer-based neural module in an online, MCTS-inspired proof search in Lean 3 and Metamath. COPRA \citep{thakur2024incontext} uses GPT-4 supplied with error feedback from the environment and lemmas from a retrieval mechanism for an agentic proof-search in Lean 3 and Coq. LLEMMA \citep{azerbayev2024llemma} continues pretraining of Code Llama on a mathematics-based corpus dubbed Proof-Pile-2, and uses their learned model for formal proof search in Lean 4. DeepSeek-Prover \cite{xin2024deepseekprover} produces synthetic Lean data en-masse for training their prover model. AlphaGeometry \citep{trinh2024solving} targets IMO problems in a geometry-specific proof assistant language using an interleaving search, where a neural module synthesizes auxiliary constructions and a symbolic engine produces deductive closures.

The Isabelle proof assistant \citep{paulson1994isabelle}, given its declarative nature and powerful symbolic automation, has too been the focus of much attention for neural theorem proving. Isabelle features Sledgehammer \citep{paulsonsledgehammer}, an automated reasoning tool which calls external automated theorem provers (ATPs) for proof synthesis. Draft, Sketch, Prove (DSP) \citep{jiang2022draft} uses a high-caliber LLM to generate natural language proofs and converts them into formal \emph{sketches} in Isabelle, whose gaps are then filled using Sledgehammer. \citet{zhao2023decomposing} employed a diffusion model to predict an optimal ordering of the few-shot examples provided to the LLM in the DSP pipeline. Lyra \citep{zheng2023lyra} utilized error-feedback from Isabelle's execution to modify holes in the sketch which were too difficult for the symbolic prover. POETRY \citep{wang2024proving} leverages recursion for theorem-proving and trains a neural module to produce proof sketches, as opposed to using in-context learning with an LLM. LEGO-Prover \citep{wang2023legoprover} extends the pipeline by incorporating a skill library which grows throughout the proof search task. Separate from approaches utilizing natural language proofs, Thor \citep{jiang2022thor} trains a transformer-based architecture to predict successful invocations of Sledgehammer, along with the usual proof-step objective. Baldur \citep{first2023baldur} explored repairing erroneous proofs in Isabelle through the use of LLMs.

The Coq interactive theorem prover has seen use in both software verification and general mathematics. Famously, mechanized proofs of the Four Colour Theorem \citep{ROBERTSON19972} and the Feit-Thompson theorem \citep{feitthompson} were produced in Coq. Similarly, numerous software verification projects have been undertaken in Coq, such as CompCert (a formally verified C compiler) and Verdi \citep{verdi} (a framework for verifying distributed systems protocols). ASTactic \citep{yang2019learning} trained a neural module involving recurrent networks and attention on data collected from various Coq repositories. 
Proverbot9001 \citep{sanchez2020generating} targeted proof synthesis on a set of held-out theorems from the CompCert project. COPRA \citep{thakur2024incontext} also evaluates on this CompCert-based task using their multi-language approach. Tactician \citep{Blaauwbroek_2020} develops a platform for proof automation for the Coq practitioner, with support for experimenting with new machine learning techniques for tactic prediction and proof search. \citep{zhang2021onlinemachinelearningtechniques} explores several online learning techniques inside Tactician, including an approximate $k$-nearest neighbors method via locality sensitive hashing which we use for our evaluation. Graph2Tac  
\citep{blaauwbroek2024graph2taconlinerepresentationlearning} uses graph neural networks for learning online hierarchical representations of new theorems and definitions, and is used for proof search within Tactician.

\section{Conclusion}
We presented \name, a benchmark for neural theorem-proving consisting of formalizations of Putnam competition problems. A distinctive feature of \name is that it spans a broad range of undergraduate-level mathematical topics, including algebra, analysis, and number theory. Another unique benefit is that it includes problems in Lean 4, Isabelle, and Coq, the three most popular formal proof frameworks.

As our experiments show, \name is a challenging benchmark: all current theorem-proving approaches fail to solve more than a handful of its problems. We believe that these failures include two root causes: (i) While current theorem-provers can effectively stitch together standard proof steps well-represented in the training corpus, they often fail at synthesizing new lemmas and orchestrating these lemmas into intricate proofs. (ii) Current methods often fail to leverage the deep knowledge available in mathematics repositories. Developing a new generation of neural theorem-provers in which these weaknesses are at least partly addressed is an exciting direction of future research.

\paragraph{Acknowledgements.}
This work was supported by NSF awards CCF-2212559 and CCF-2403211, the NSF Institute for Foundations of Machine Learning, and a gift by the Aziz Family Foundation. We thank Oliver Nash, Eric Wieser, Edward Lockhart, Fabian Gloeckle, Karl Palmskog, Lasse Blaauwbroek, Jason Rute, and Kaiyu Yang for useful discussions, aiding in benchmark maintenance, and support with setting up experiments.

\bibliographystyle{neurips_conference}
\bibliography{neurips_conference}

\newpage

\section*{Checklist}

\begin{enumerate}

\item For all authors...
\begin{enumerate}
  \item Do the main claims made in the abstract and introduction accurately reflect the paper's contributions and scope?
    \answerYes{} We support our main claims in \Cref{sec:benchmark} and \Cref{sec:eval}.
  \item Did you describe the limitations of your work?
    \answerYes{} We discussed in \Cref{sec:benchmark} the challenges of formalizing certain problem categories such as geometry and probability due to the nature of support for such mathematical theory in each language.
  \item Did you discuss any potential negative societal impacts of your work?
    \answerNA{} We do not anticipate any negative societal impact of our work.
  \item Have you read the ethics review guidelines and ensured that your paper conforms to them?
    \answerYes{} We have read the ethics review guidelines and ensured our paper conforms to them.
\end{enumerate}

\item If you are including theoretical results...
\begin{enumerate}
  \item Did you state the full set of assumptions of all theoretical results?
    \answerNA{} We do not include any theoretical results.
	\item Did you include complete proofs of all theoretical results?
    \answerNA{} We do not include any theoretical results.
\end{enumerate}

\item If you ran experiments (e.g. for benchmarks)...
\begin{enumerate}
  \item Did you include the code, data, and instructions needed to reproduce the main experimental results (either in the supplemental material or as a URL)?
    \answerYes{} We disclosed all information related to the experiments, which use open-sourced methods. We have also included the URL to our dataset: \href{https://github.com/trishullab/PUTNAM/}{https://github.com/trishullab/PUTNAM/}.
  \item Did you specify all the training details (e.g., data splits, hyperparameters, how they were chosen)?
    \answerNA{} We did not perform any training.
	\item Did you report error bars (e.g., with respect to the random seed after running experiments multiple times)?
    \answerNo{} We evaluate our selected methodologies using established metrics accepted by the neural theorem-proving community. See \Cref{sec:eval}.
	\item Did you include the total amount of compute and the type of resources used (e.g., type of GPUs, internal cluster, or cloud provider)? 
    \answerYes{} Most of our experiments rely on calls to GPT-4, we include sampling details. We also mention the hyperparameters to calls to symbolic methods in \Cref{sec:eval}.
\end{enumerate}

\item If you are using existing assets (e.g., code, data, models) or curating/releasing new assets...
\begin{enumerate}
  \item If your work uses existing assets, did you cite the creators?
    \answerYes{} We did cite the creators of any existing assets we used.
  \item Did you mention the license of the assets?
    \answerYes{} We aligned the license of our benchmark with the license of those assets.
  \item Did you include any new assets either in the supplemental material or as a URL?
    \answerYes{} We included our dataset by sharing the following URL: \href{https://github.com/trishullab/PUTNAM/}{https://github.com/trishullab/PUTNAM/}.
  \item Did you discuss whether and how consent was obtained from people whose data you're using/curating? 
    \answerYes{} We obtained permission from the MAA.
  \item Did you discuss whether the data you are using/curating contains personally identifiable information or offensive content?
    \answerNA{} Our data does not contain such content.
\end{enumerate}

\item If you used crowdsourcing or conducted research with human subjects...
\begin{enumerate}
  \item Did you include the full text of instructions given to participants and screenshots, if applicable?
    \answerNA{} We did not conduct research with human subjects nor crowdsource.
  \item Did you describe any potential participant risks, with links to Institutional Review Board (IRB) approvals, if applicable?
    \answerNA{} We did not conduct research with human subjects nor crowdsource.
  \item Did you include the estimated hourly wage paid to participants and the total amount spent on participant compensation?
    \answerNA{} We did not conduct research with human subjects nor crowdsource.
\end{enumerate}

\end{enumerate}


\newpage
\appendix
\section{Appendix}
We include further examples of formalizations from \name below.
\begin{figure}[H]
\begin{mdframed}[roundcorner=10pt]
\begin{lstlisting}
From mathcomp Require Import ssrbool seq ssrnat prime rat ssralg ssrnum ssrint.

Local Open Scope ring_scope.

Theorem putnam_2009_b1 :
  let fact_prod (ls : seq nat) : rat := \prod_(i <- ls) (i`!)%:Q in
  forall q : rat, q > 0 -> exists n d : seq nat,
  all prime (n ++ d) /\ fact_prod n / fact_prod d = q.
Proof. Admitted.
\end{lstlisting}
\end{mdframed}
\caption{A formalization of Putnam 2009 B1 in Coq relying on the MathComp repository.} 
\label{fig:coq-formalization-cast-example}
\end{figure}

\begin{figure}[H]
\begin{mdframed}[roundcorner=10pt]
\textbf{Putnam 2001 B4.} Let $S$ denote the set of rational numbers different from $\{-1, 0, 1\}$. Define $f : S \to S$ by $f(x) = x - 1/x$. Prove or disprove that
\[
\bigcap_{n = 1}^\infty f^{(n)}(S) = \varnothing,
\]
where $f^{(n)}$ denotes $f$ composed with itself $n$ times.
\end{mdframed}
\begin{mdframed}[roundcorner=10pt]
\begin{lstlisting}
abbrev putnam_2001_b4_solution : Prop := True
theorem putnam_2001_b4
    (S : Set ℚ)
    (hS : S = univ \ {-1, 0, 1})
    (f : S → S)
    (hf : ∀ x : S, f x = x - 1 / (x : ℚ))
    : ⋂ n ∈ Ici 1, f^[n] '' univ = ∅ ↔ putnam_2001_b4_solution 
    := sorry
\end{lstlisting}
\end{mdframed}
\caption{A formalization of Putnam 2001 B4 in Lean 4. As the problem requires deciding whether the infinite intersection is empty, it is not directly the statement of a theorem. We consider the associated ``solution'' of this problem to be a boolean value, and factor it out from the theorem statement. \texttt{sorry} is the placeholder keyword for Lean.} \label{fig:solution-prop-example} 
\end{figure}
\begin{figure}
\begin{mdframed}[roundcorner=10pt]
\textbf{Putnam 2020 A3.} Let $a_0 = \pi/2$, and let $a_n = \sin(a_{n-1})$ for $n \geq 1$. Determine whether 
\[
\sum_{n = 1}^\infty a_n^2
\]
converges.
\end{mdframed}
\begin{mdframed}[roundcorner=10pt]
\begin{lstlisting}
abbrev putnam_2020_a3_solution : Prop := False
theorem putnam_2020_a3
    (a : ℕ → ℝ)
    (ha0 : a 0 = Real.pi / 2)
    (ha : ∀ n : ℕ, n ≥ 1 → a n = Real.sin (a (n - 1)))
    : (∃ L : ℝ, Tendsto (fun m : ℕ => ∑ n : Icc 1 m, (a n)^2) atTop (N L)) 
        ↔ putnam_2020_a3_solution 
    := sorry

\end{lstlisting}
\end{mdframed}
\caption{A formalization of Putnam 2020 A3 in Lean 4. As the problem requires deciding whether the series converges, it is not directly the statement of a theorem. We consider the associated ``solution'' of this problem to be a boolean value, and factor it out from the theorem statement.} \label{fig:solution-prop-example-2} 
\end{figure}

\begin{figure}
\begin{mdframed}[roundcorner=10pt]
\textbf{Putnam 1997 A4.} Let $G$ be a group with identity $e$ and $\phi : G \to G$ a function such that
\[
\phi(g_1)\phi(g_2)\phi(g_3) = \phi(h_1)\phi(h_2)\phi(h_3)
\]
whenever $g_1g_2g_3 = e = h_1h_2h_3$. Prove that there exists an element $a \in G$ such that $\psi(x) = a\phi(x)$ is a homomorphism.
\end{mdframed}
\begin{mdframed}[roundcorner=10pt]
\begin{lstlisting}
theorem putnam_1997_a4
    (G : Type*)
    [Group G]
    (φ : G → G)
    (hφ : ∀ g1 g2 g3 h1 h2 h3 : G, (g1 * g2 * g3 = 1 ∧ h1 * h2 * h3 = 1) 
    → φ g1 * φ g2 * φ g3 = φ h1 * φ h2 * φ h3)
    : ∃ a : G, let ψ := fun g => a * φ g; ∀ x y : G, ψ (x * y) = ψ x * ψ y 
    := sorry

\end{lstlisting}
\end{mdframed}
\caption{A formalization of Putnam 1997 A4, which requires knowledge of group theory, in Lean 4. The informal statement is slightly underspecified - $g_1, g_2, g_3, h_1, h_2, h_3$ are not explicitly defined to be in $G$. To produce the formalization, we must be specific about the type of $g_i, h_i$.} \label{fig:group-theory-example} 
\end{figure}

\begin{figure}
\begin{mdframed}[roundcorner=10pt]
\textbf{Putnam 2018 B1.} Let $\mathcal{P}$ be the set of vectors defined by 
\[
\mathcal{P}=\left\{\left.\begin{pmatrix} a \\ b \end{pmatrix}\right| 0 \leq a \leq 2, 0 \leq b \leq 100,\text{ and }a,b \in \mathbb{Z}\right\}
\]
Find all $\mathbf{v} \in \mathcal{P}$ such that the set $\mathcal{P} \setminus \{\mathbf{v}\}$ obtained by omitting vector $\mathbf{v}$ from $\mathcal{P}$ can be partitioned into two sets of equal size and equal sum.
\end{mdframed}
\begin{mdframed}[roundcorner=10pt]
\begin{lstlisting}
abbrev putnam_2018_b1_solution : Set (Vector ℤ 2) := 
    {v : Vector ℤ 2 | ∃ b : ℤ, 0 ≤ b ∧ b ≤ 100 ∧ Even b ∧ v.toList = [1, b]}
theorem putnam_2018_b1
(v : Mathlib.Vector ℤ 2)
(P Pvdiff : Finset (Mathlib.Vector ℤ 2))
(hP : P = 
    {v' : Mathlib.Vector ℤ 2 | 0 ≤ v'[0] ∧ v'[0] ≤ 2 ∧ 0 ≤ v'[1] ∧ v'[1] ≤ 100})
(hPvdiff : Pvdiff = P \ ({v} : Finset (Mathlib.Vector ℤ 2)))
: (v ∈ P ∧ (∃ Q R : Finset (Mathlib.Vector ℤ 2),
    (Q ∪ R = Pvdiff) ∧ (Q ∩ R = ∅) ∧ (Q.card = R.card) ∧
    (∑ q in Q, q[0] = ∑ r in R, r[0]) ∧ (∑ q in Q, q[1] = ∑ r in R, r[1])))
  ↔ v ∈ putnam_2018_b1_solution :=
sorry
\end{lstlisting}
\end{mdframed}
\caption{A formalization of Putnam 2018 B1, which requires the Vector class from mathlib4.} \label{fig:vector-example} 
\end{figure}

\begin{figure}
\begin{mdframed}[roundcorner=10pt]
\textbf{Putnam 1992 B6.} Let $\mathcal{M}$ be a set of real $n \times n$ matrices such that
\begin{enumerate}
    \item $I \in \mathcal{M}$, where $I$ is the $n \times n$ identity matrix;
    \item if $A \in \mathcal{M}$ and $B \in \mathcal{M}$, then exactly one of $AB \in \mathcal{M}$ and $-AB \in \mathcal{M}$ holds;
    \item if $A \in \mathcal{M}$ and $B \in \mathcal{M}$, then either $AB = BA$ or $AB = -BA$;
    \item if $A \in \mathcal{M}$ and $A \neq I$, there is at least one $B \in \mathcal{M}$ such that $AB = -BA$.
\end{enumerate}
Prove that $\mathcal{M}$ contains at most $n^2$ matrices.
\end{mdframed}
\begin{mdframed}[roundcorner=10pt]
\begin{lstlisting}
theorem putnam_1992_b6:
  fixes n :: nat
    and M :: "(real^'n^'n) set"
  assumes npos: "n > 0"
    and pncard: "CARD('n) = n"
    and h1: "mat 1 ∈ M"
    and h2: "∀A∈M. ∀B∈M. (A**B ∈ M) ≠ (-A**B ∈ M)"
    and h3: "∀A∈M. ∀B∈M. (A**B = B**A) ∨ (A**B = -B**A)"
    and h4: "∀A∈M. (A ≠ mat 1 → (∃B∈M. A**B = -B**A))"
  shows "card M ≤ n^2"
  sorry
\end{lstlisting}
\end{mdframed}
\caption{An Isabelle formalization of Putnam 1992 B6.} \label{fig:isabelle-formalization-example} 
\end{figure}

\begin{figure}
\begin{mdframed}[roundcorner=10pt]
\textbf{Putnam 2012 A3.} Let $f : [-1,1] \to \mathbb{R}$ be a continuous function such that
\begin{enumerate}
    \item $f(x) = \frac{2-x^2}{2}f(\frac{x^2}{2-x^2})$ for every $x$ in $[-1, 1]$,
    \item $f(0) = 1,$ and
    \item $\lim_{x \to 1^-} \frac{f(x)}{\sqrt{1-x}}$ exists and is finite.
\end{enumerate}
Prove that $f$ is unique, and express $f(x)$ in closed form.
\end{mdframed}
\begin{mdframed}[roundcorner=10pt]
\begin{lstlisting}
definition putnam_2012_a3_solution :: "real ⇒ real" where 
  "putnam_2012_a3_solution ≡ (λx::real. sqrt (1 - x^2))"
theorem putnam_2012_a3:
  fixes S :: "real set"
  and hf :: "(real ⇒ real) ⇒ bool"
  defines "S ≡ {-1..1}"
  and "hf ≡ (λf::real⇒real. continuous_on S f ∧
    (∀x∈S. f x = ((2 - x^2)/2)*f (x^2/(2 - x^2))) ∧ f 0 = 1 ∧
    (∃y::real. filterlim (λx::real. (f x)/sqrt (1 - x)) (nhds y) (at_left 1)))"
  shows "hf putnam_2012_a3_solution ∧ 
    (∀f::real⇒real. hf f → (∀x∈S. f x = putnam_2012_a3_solution x))"
  sorry
\end{lstlisting}
\end{mdframed}
\caption{An Isabelle formalization of Putnam 2012 A3. The mechanism for factoring the solution out of the theorem statement is similar to that of Lean.} \label{fig:solution-func-isabelle-example} 
\end{figure}

\begin{figure}
\begin{mdframed}[roundcorner=10pt]
\textbf{Putnam 1980 A5.} Let $P(t)$ be a nonconstant polynomial with real coefficients. Prove that the system of simultaneous equations
\[
0 = \int_0^x P(t) \sin t dt = \int_0^x P(t) \cos t dt
\]
has only finitely many real solutions $x$.
\end{mdframed}
\begin{mdframed}[roundcorner=10pt]
\begin{lstlisting}
From mathcomp Require Import all_algebra all_ssreflect.
From mathcomp Require Import reals trigo lebesgue_integral lebesgue_measure measure.
From mathcomp Require Import classical_sets cardinality.

Set Implicit Arguments.
Unset Strict Implicit.
Unset Printing Implicit Defensive.

Local Open Scope classical_set_scope.
Local Open Scope ring_scope.

Variable R : realType.
Definition mu := [the measure _ _ of @lebesgue_measure R].
Theorem putnam_1980_a5
    (P : {poly R})
    (Pnonconst : gtn (size P) (1%nat))
    : finite_set [set x : R |
    \int[mu]_(t in [set t : R | 0 <= t <= x]) (fun y => P.[y] * (sin y)) t = 0 /\ 
    \int[mu]_(t in [set t : R | 0 <= t <= x]) (fun y => P.[y] * (cos y)) t = 0].
Proof. Admitted.
\end{lstlisting}
\end{mdframed}
\caption{A Coq formalization of Putnam 1980 A5. This formalization is done using Coquelicot, a Coq repository outside of the standard library. The Coq equivalent of \texttt{sorry} is \texttt{Admitted}.} \label{fig:coq-formalization-example} 
\end{figure}

\begin{figure}
\begin{mdframed}[roundcorner=10pt]
\textbf{Putnam 2017 B2.} Suppose that a positive integer $N$ can be expressed as the sum of $k$ consecutive positive integers
\[
N = a + (a + 1) + (a + 2) + \dots + (a + k - 1)
\]
for $k = 2017$ but for no other values of $k > 1$. Considering all positive integers $N$ with this property, what is the smallest positive integer $a$ that occurs in any of these expressions?
\end{mdframed}
\begin{mdframed}[roundcorner=10pt]
\begin{lstlisting}
From mathcomp Require Import all_ssreflect all_algebra.

Set Implicit Arguments.
Unset Strict Implicit.
Unset Printing Implicit Defensive.

Local Open Scope ring_scope.

Definition putnam_2017_b2_solution : nat := 16.
Theorem putnam_2017_b2 :
    let seq (a : int) (k : nat) := \sum_(0 <= i < k) (a + i%:Z) in
    let valid (a : int) := a > 0 /\ (forall (b : int) (k : nat), b > 0 -> gt k 1 -> seq a 2017%nat = seq b k -> k = 2017%nat) in
    valid putnam_2017_b2_solution /\ (forall a, valid a -> a >= putnam_2017_b2_solution%:Z).
Proof. Admitted.
\end{lstlisting}
\end{mdframed}
\caption{A Coq formalization of Putnam 2017 B2. As the problem requires a numerical witness, we factor that out using Coq's syntax for making definitions.} \label{fig:coq-solution-example} 
\end{figure}

\begin{figure}
    
\end{figure}

\begin{figure}
\begin{mdframed}[roundcorner=10pt]
\textbf{Putnam 1988 B1.} A \emph{composite} is a product $ab$ with $a$ and $b$ not necessarily distinct integers $\{2,3,4,\dots\}$. Show that every composite is expressible as $xy + xz + yz + 1$ with $x,y,z$ positive integers.
\end{mdframed}
\begin{mdframed}[roundcorner=10pt]
\begin{lstlisting}
Require Import ZArith Znumtheory Lia.
Open Scope Z.
Theorem putnam_1988_b1:
    forall (a : Z), a >= 2 -> 
    forall (b : Z), b >= 2 -> 
    exists (x y z: Z), x > 0 /\ y > 0 /\ z > 0 /\ 
    a * b = x * y + y * z + z * x + 1.
Proof.
    intros a Ha b Hb.
    exists 1, (a - 1), (b - 1).
    split.
    - lia.
    - split.
    + lia.
    + split.
    * lia.
Qed.

\end{lstlisting}
\end{mdframed}
\caption{A Coq proof of Putnam 1988 B1 generated through a few-shot invocation of GPT-4. The proof is similar to that of the Lean version, also discovered by GPT-4. The main difficulty of the problem is to choose the values of $x,y,z$ given $a,b$. Once correctly supplied, the remainder of the proof is routine and can be done with automated methods like \texttt{lia} which handles linear arithmetic.} \label{fig:coq-proof} 
\end{figure}

\begin{figure}
\begin{minipage}{0.5\textwidth}
\begin{mdframed}[roundcorner=10pt]
\begin{lstlisting}
theorem mathd_numbertheory_85 :
  1 * 3^3 + 2 * 3^2 + 2*3 + 2 = 53 
  := sorry
\end{lstlisting}
\end{mdframed}

\label{fig:easy-miniF2F-Lean}
\end{minipage}
\hspace{0.1in}
\begin{minipage}{0.47\textwidth}

\begin{mdframed}[roundcorner=10pt]
\begin{lstlisting}
theorem mathd_algebra_107
(x y : ℝ)
(h₀ : x^2 + 8 * x + y^2 - 6 * y = 0) 
: (x + 4)^2 + (y-3)^2 = 5^2 := sorry
\end{lstlisting}
\end{mdframed}

\label{fig:easy-miniF2F-Lean-2}
\end{minipage}
\caption{Examples of formalizations of easy problems in \minif. While useful for benchmarking straightforward mathematical reasoning in a formal setting, these problems are quite simple compared to the competition problems present in \name. We note that \minif does include some formalizations of problems sourced directly from high school competitions, but these are fewer in number.} 
\vspace{-0.25in}
\end{figure}

\begin{figure}
\begin{mdframed}[roundcorner=10pt]
\begin{minted}[breaklines]{md}
You are proficient at formal theorem-proving in Lean 4. Given a theorem statement in Lean 4, generate the proof in Lean 4. You can assume that you have access to Lean's mathlib library.

The theorem is described in the following format:
1. The theorem statement using the `[THEOREM]` keyword.
3. The theorem description ends with the keyword `[END]`.

Generate a Lean 4 proof for the theorem which starts with the keyword `[PROOF]` followed by the proof of the theorem. The syntax for Lean 4 is different than that of Lean 3 - premises like "Nat.dvd_mul" and "Finset.singleton_injective" exist in Lean 4, the equivalent in Lean 3 is "nat.dvd_mul" and "finset.singleton_injective" which DO NOT WORK in Lean 4. Additionally, you cannot chain tactics into one step using ',' - this will NOT work - you can use ';' instead but try to avoid such usage where not necessary! When doing rewrites you MUST wrap the premise in brackets: "rw [h]". If you want to do multiple rewrites at once you can do something like "rw [step1, step2, step3]". Always predict one tactic at a time, though you can predict the "have" tactic and may supply a proof for it with tactics split by ";". You can provide witnesses to consecutive existential quantifiers all at once, for example 'use 1, 2, 3' but NOT as a list 'use [1, 2, 3]' - these are not the same things!  You can introduce with "intro" everything you think you can introduce at once. In Lean 4, you can split apart conjunctions with "constructor" NOT "split". You should use the "ring" tactic to handle goals that follow from ring axioms, especially instead of doing a long series of rewrites or calculations. Similarly, "linarith" can be useful for solving goals involving linear arithmetic. Do NOT indent tactics, every new line should not have spaces to start! PLEASE use Lean 4 syntax only! The proof ends with the keyword `[END]`. Also please DO NOT write `sorry` in the proof. You can assume that the theorem is provable.
\end{minted}
\end{mdframed}
\caption{Parts of the ``system prompt'' used by GPT-4 for Lean 
4 evaluations. Due to GPT-4's tendency towards producing outputs in Lean 3 syntax, our prompt places special attention towards preventing such syntactic mistakes. A similar modification is made to COPRA's system prompt for Lean 3.}
\label{fig:gpt-prompt}
\end{figure}

\begin{figure}
\begin{mdframed}[roundcorner=10pt]
\begin{minted}[breaklines]{md}
Goals to prove:
[GOALS]
[GOAL] 1
DifferentiableAt ℝ (fun x => g x / hg0 x) 0 → DifferentiableAt ℝ g 0
[HYPOTHESES] 1
[HYPOTHESIS] case mpr
[HYPOTHESIS] f : True
[HYPOTHESIS] g hg0 : ℝ → ℝ
[HYPOTHESIS] hcg : hg0 0 ≠ 0
[HYPOTHESIS] hfg : ContinuousAt hg0 0
[HYPOTHESIS] hfg_div : DifferentiableAt ℝ (fun x => g x * hg0 x) 0

[STEPS]
[STEP] constructor
[STEP] intro h
[STEP] trivial
[STEP] intros f g hg0 hcg hfg hfg_div

[INCORRECT STEPS]
[STEP] apply differentiable_at_of_mul

[LAST STEP]
apply differentiable_at.div

[ERROR MESSAGE]
error: unknown identifier 'differentiable_at.div'
[END]
\end{minted}
\end{mdframed}
\caption{An example of a failed tactic prediction during proof search for Putnam 2011 B2 using COPRA in Lean 4. GPT-4 predicts a tactic involving the premise ``differentiable\_at.div,'' which exists in Lean 3, but not Lean 4. Even with the system prompt asserting outputs should involve Lean 4 syntax alone, GPT-4 is not always capable of making the distinction.}
\label{fig:appendix-copra-lean3-failure}
\end{figure}

\begin{figure}
\begin{mdframed}[roundcorner=10pt]
\begin{lstlisting}
theorem putnam_2001_a1
(S : Type*)
[Mul S]
(hS : ∀ a b : S, (a * b) * a = b)
: ∀ a b : S, a * (b * a) = b := by
intro a b
have h1 : (a * (b * a)) * a = b * a := by
 rw [←mul_assoc]
 rw [hS]
 rw [hS]
 apply h1
\end{lstlisting}
\end{mdframed}
\caption{A failed proof generated by few-shot invocation of GPT-4. GPT-4 misunderstands that the hypothesis \texttt{[Mul S]}, which gives an operation $\star$ and asserts it is a binary operation on $S$, also asserts associativity of the operation. The tactic \texttt{rw [←mul\_assoc]}, which performs a rewrite using the associativity of $\star$, is hence not applicable in this setting. } \label{fig:appendix-copra-premise-failure} 
\end{figure}

\begin{figure}
\begin{mdframed}[roundcorner=10pt]
\textbf{Putnam 1986 B1.} Inscribe a rectangle of base $b$ and height $h$ in a circle of radius one, and inscribe an isosceles triangle in the region of the circle cut off by one base of the rectangle. For what values of $h$ do the rectangle and triangle have the same area?
\end{mdframed}
\begin{mdframed}[roundcorner=10pt]
\begin{lstlisting}
theorem putnam_1986_b1:
fixes b h :: "real"
assumes hbh : "b > 0 ∧ h > 0 ∧ b^2 + h^2 = 2^2"
and areaeq : "b*h = 0.5 * b * (1 - h/2)"
shows "h = ( 2 / 5 )"
proof - 
from assms have eq1: "b^2 + h^2 = 4" by simp
  from areaeq have eq2: "b * h = 0.5 * b * (1 - h/2)" by simp
  then have eq3: "b * h = 0.5 * b - 0.25 * b * h" by algebra
  then have eq4: "1.25 * b * h = 0.5 * b" by algebra
  then have eq5: "h = 0.5 / 1.25" using assms(1) by simp
  then show ?thesis by simp
qed
\end{lstlisting}
\end{mdframed}
\caption{A successful proof in Isabelle discovered using GPT-4. While the theorem statement differs from the formalization, which is algebraic, we note that the official solution is also algebraic and is similar to the generated Isabelle proof.} \label{fig:geometric-problem} 
\end{figure}

\begin{figure}
\begin{mdframed}[roundcorner=10pt]
\textbf{Putnam 1995 A1.} Let $S$ be a set of real numbers which is closed under multiplication (that is, if $a$ and $b$ are in $S$, then so is $ab$. Let $T$ and $U$ be disjoint subsets of $S$ whose union is $S$. Given that the product of any \emph{three} (not necessarily distinct) elements of $T$ is in $T$ and the product of any three elements of $U$ is in $U$, show that at least one of the two subsets $T, U$ is closed under multiplication.
\end{mdframed}
\begin{mdframed}[roundcorner=10pt]
\begin{lstlisting}
theorem putnam_1995_a1:
  fixes S :: "real set"
    and T U :: "real set"
  assumes hS: "∀a∈S. ∀b∈S. a * b ∈ S"
    and hsub: "T ⊆ S ∧ U ⊆ S"
    and hunion: "(T ∪ U) = S"
    and hdisj: "(T ∩ U) = {}"
    and hT3: "∀a∈T. ∀b∈T. ∀c∈T. a * b * c ∈ T"
    and hU3: "∀a∈U. ∀b∈U. ∀c∈U. a * b * c ∈ U"
  shows "(∀a∈T. ∀b∈T. a * b ∈ T) ∨ (∀a∈U. ∀b∈U. a * b ∈ U)"
proof (rule ccontr)   
  assume "¬ ((∀a∈T. ∀b∈T. a * b ∈ T) ∨ (∀a∈U. ∀b∈U. a * b ∈ U))"   
  then have "∃t1∈T. ∃t2∈T. t1 * t2 ∉ T" "∃u1∈U. ∃u2∈U. u1 * u2 ∉ U" by auto
  then obtain t1 t2 where t1t2: "t1 ∈ T" "t2 ∈ T" "t1 * t2 ∈ U" 
    using hS hunion by blast  
  then obtain u1 u2 where u1u2: "u1 ∈ U" "u2 ∈ U" "u1 * u2 ∈ T" 
    using ‹¬ ((∀a∈T. ∀b∈T. a * b ∈ T) ∨ (∀a∈U. ∀b∈U. a * b ∈ U))› hS hunion
    by blast
  have "t1 * t2 * t3 ∈ T" if "t3 ∈ T" for t3 using hT3 t1t2(1) t1t2(2) that 
    by auto
  then have "t1 * t2 * t3 ∈ U" if "t3 ∈ T" for t3 using t1t2(3) hS that
    by (metis disjoint_iff hU3 hdisj mult.commute u1u2(1) u1u2(2) u1u2(3))
  then have False using hdisj t1t2(1) t1t2(2)
    by (meson disjoint_iff_not_equal hT3) 
  then show False by blast
qed
\end{lstlisting}
\end{mdframed}
\caption{A successful proof in Isabelle discovered using DSP. In particular, this proof is not found using a single invocation of sledgehammer, so the sketching mechanism of DSP is crucial for this problem. We note that the DSP pipeline involves using an LLM (GPT-4) to synthesize an informal proof which is translated into a sketch in Isabelle - this can potentially be a source of indirect dataset contamination, as we cannot ensure the informal proofs are not present in GPT-4's training data.} \label{fig:dsp-example} 
\end{figure}

\begin{figure}
\begin{mdframed}[roundcorner=10pt]
\begin{lstlisting}
theorem putnam_1971_b1:
  fixes Smul :: "'S ⇒ 'S ⇒ 'S" (infixl "*" 70)
  assumes hself: "∀x::'S. x * x = x"
    and h2: "∀x y z::'S. (x * y) * z = (y * z) * x"
  shows "∀x y z::'S. (x * y) *  z = x * (y * z) ∧ x * y = y * x"
proof -
  have comm: "∀x y::'S. x * y = y * x"
  proof
    fix x y :: 'S
    have "(x * y) * x = (y * x) * x" using h2 by blast (* sledgehammer *)
    also have "... = y * x" using hself by (metis h2) (* sledgehammer *)
    finally have "(x * y) * x = y * x" by simp (* sledgehammer *)
    then have "x * y = y * x" using hself by (metis h2) (* sledgehammer *)
    thus "x * y = y * x" by simp
  qed
  have assoc: "∀x y z::'S. (x * y) * z = x * (y * z)"
  proof
    fix x y z :: 'S
    have "(x * y) * z = (y * z) * x" using h2 sledgehammer
    also have "... = x * (y * z)" using comm sledgehammer
    finally show "(x * y) * z = x * (y * z)" by simp
  qed
  show "∀x y z::'S. (x * y) * z = x * (y * z) ∧ x * y = y * x"
    using assoc comm by simp
qed

\end{lstlisting}
\end{mdframed}
\caption{An erroneous generation produced by DSP for Putnam 1971 B1. While a single invocation of Sledgehammer can prove this problem, the formal sketch generated in the pipeline is much more complex and is erroneous, leading to a failed proof attempt.} \label{fig:dsp-fail-example} 
\end{figure}

\begin{figure}
\begin{mdframed}[roundcorner=10pt]
\begin{lstlisting}
theorem putnam_1971_b1:
  fixes Smul :: "'S ⇒ 'S ⇒ 'S" (infixl "*" 70)
  assumes hself: "∀x::'S. x * x = x"
  and h2: "∀x y z::'S. (x * y) * z = (y * z) * x"
  shows "∀x y z::'S. (x * y) * z = x * (y * z) ∧ x * y = y * x"
\end{lstlisting}
\end{mdframed}
\begin{mdframed}[roundcorner=10pt]
\begin{lstlisting}
theorem putnam_2012_a2:
  fixes Smul :: "'S ⇒'S ⇒ 'S" (infixl "*" 70)
  and a b c :: 'S
  assumes Smulasg: "abel_semigroup Smul"
  and hS: "∀x y::'S. ∃z::'S. x * z = y"
  and habc: "a * c = b * c"
shows "a = b"
\end{lstlisting}
\end{mdframed}
\caption{The other two Isabelle formalizations solved with invocations of Sledgehammer. We note that the problems Sledgehammer was capable of solving are all problems involving binary operations on sets. It is not surprising that SMT solvers are capable of solving such problems, which do not require reasoning about complicated objects. } 
\label{fig:isabelle-sledgehammer-others} 
\end{figure}

\begin{figure}
\begin{mdframed}[roundcorner=10pt]
\begin{minted}[breaklines]{md}
Goals to prove:
[GOALS]
[GOAL] 1
1 > 0 /\
1 > 0 /\
a * b - 1 > 0 /\ a * b = 1 * 1 + 1 * (a * b - 1) + (a * b - 1) * 1 + 1
[HYPOTHESES] 1
[HYPOTHESIS] a : Z
[HYPOTHESIS] Ha : a >= 2
[HYPOTHESIS] b : Z
[HYPOTHESIS] Hb : b >= 2

[STEPS]
[STEP] intros a Ha b Hb.

[LAST STEP]
exists 1, 1, (a * b - 1).

[SUCCESS]
[END]
...
Goals to prove: # Step 32 of Search
[GOALS]
[GOAL] 1
a - 1 > 0 /\
1 > 0 /\
b - 1 > 0 /\ a * b = (a - 1) * 1 + 1 * (b - 1) + (b - 1) * (a - 1) + 1
[HYPOTHESES] 1
[HYPOTHESIS] a : Z
[HYPOTHESIS] Ha : a >= 2
[HYPOTHESIS] b : Z
[HYPOTHESIS] Hb : b >= 2

[STEPS]
[STEP] intros a Ha b Hb.

[LAST STEP]
exists (a - 1), 1, (b - 1).

[SUCCESS]
[END]
\end{minted}
\end{mdframed}
\caption{Early in COPRA's attempt on Putnam 1988 B1, an incorrect prediction of $x,y,z$ given $a,b$ is made, which dooms that path of search as the most crucial step is the correct choice. Later, at step 32 of search, COPRA backtracks and then successfully predicts a correct choice for $x,y,z$. Once this step is generated, the remainder of the proof is straightforward. }
\label{fig:appendix-copra-lean4-failure-success}
\end{figure}

\end{document}